\def\year{2020}\relax
\documentclass[letterpaper]{article} 
\usepackage{aaai20}  
\usepackage{times}  
\usepackage{helvet} 
\usepackage{courier}  
\usepackage[hyphens]{url}  
\usepackage{graphicx} 
\usepackage{subfigure}
\usepackage{amsmath}
\usepackage{float}
\usepackage{multirow}
\usepackage{enumerate}
\usepackage{enumitem} 
\usepackage[T2A,T1]{fontenc}
\usepackage[utf8]{inputenc}
\usepackage[russian,english]{babel}
\usepackage[normalem]{ulem}
\usepackage{makecell}
\usepackage{booktabs}

\usepackage{graphicx}  
\newcommand{\citet}[1]{\citeauthor{#1}~\shortcite{#1}}

\begin{document}
%
\title{
Cross-lingual Pre-training Based Transfer for Zero-shot Neural \\ Machine Translation}


\author{ Baijun Ji$^\ddag$,  Zhirui Zhang$^\S$, Xiangyu Duan$^{\dag\ddag}$\thanks{Corresponding Author.}, Min Zhang$^{\dag\ddag}$, Boxing Chen$^\S$ and Weihua Luo$^\S$ \\
  $^\dag$Institute of Artificial Intelligence, Soochow University, Suzhou, China \\
  $^\ddag$School of Computer Science and Technology, Soochow University, Suzhou, China \\
  $^\S$Alibaba DAMO Academy, Hangzhou, China \\
  $^{\ddag}$bjji@stu.suda.edu.cn \quad $^{\dag}$\{xiangyuduan, minzhang\}@suda.edu.cn \\
  $^\S$\{zhirui.zzr,boxing.cbx,weihua.luowh\}@alibaba-inc.com\  
}

\maketitle
\begin{abstract}

Transfer learning between different language pairs has shown its effectiveness for Neural Machine Translation (NMT) in low-resource scenario.
However, existing transfer methods involving a common target language are far from success in the extreme scenario of zero-shot translation, due to the language space mismatch problem between transferor (the parent model) and transferee (the child model) on the source side.
To address this challenge, we propose an effective transfer learning approach based on cross-lingual pre-training.
Our key idea is to make all source languages share the same feature space and thus enable a smooth transition for zero-shot translation.
To this end, we introduce one monolingual pre-training method and two bilingual pre-training methods to obtain a universal encoder for different languages.
Once the universal encoder is constructed, the parent model built on such encoder is trained with large-scale annotated data and then directly applied in zero-shot translation scenario.
Experiments on two public datasets show that our approach significantly outperforms strong pivot-based baseline and various multilingual NMT approaches.

\end{abstract}

\section{Introduction}

\begin{figure}
\centering
\includegraphics[width=1\linewidth,height=30mm]{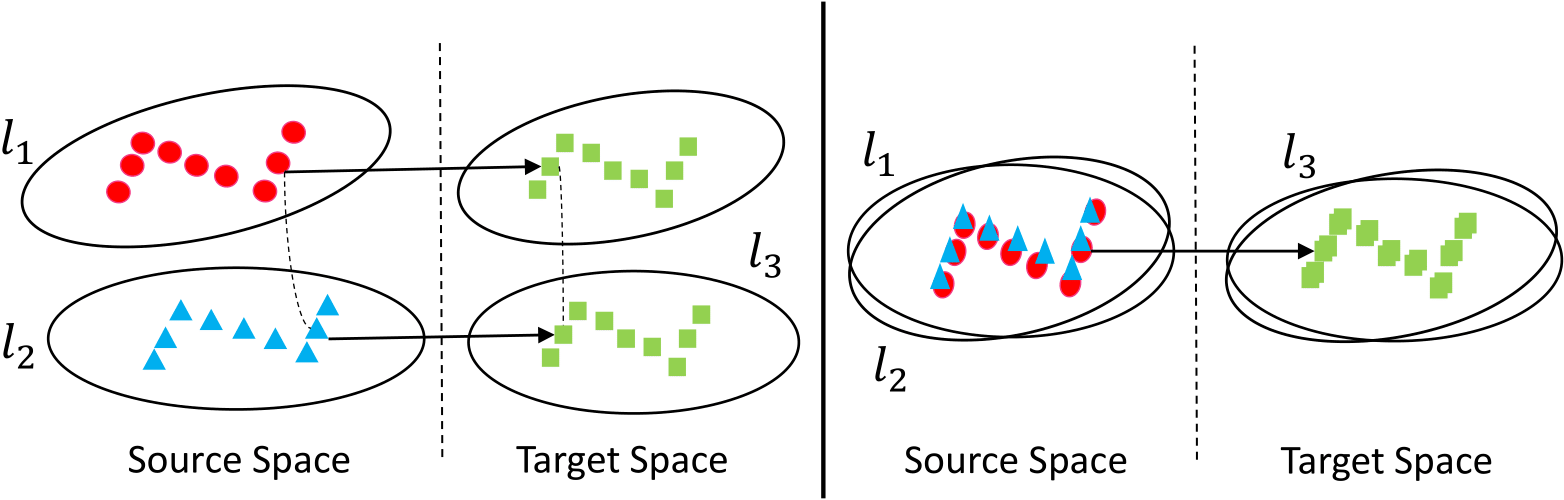}
\caption{The circle and triangle dots represent source sentences in different language $l_1$ and $l_2$, and the square dots means target sentences in language $l_3$. A sample of translation pairs is connected by the dashed line. We would like to force each of the translation pairs has the same latent representation as the right part of the figure so as to transfer  $l_1 \to l_3$ model directly to $l_2 \to l_3$ model.} 
\label{fig:intro-space-map}
\end{figure}


Although Neural Machine Translation (NMT) has dominated recent research on translation tasks~\cite{wu2016google,vaswani2017attention,hassan2018achieving}, NMT heavily relies on large-scale parallel data, resulting in poor performance on low-resource or zero-resource language pairs~\cite{koehn2017six}.
Translation between these low-resource languages (e.g., Arabic$\to$Spanish) is usually accomplished with pivoting through a rich-resource language (such as English), i.e., Arabic (source) sentence is translated to English (pivot) first which is later translated to Spanish (target)~\cite{kauers2002interlingua,de2006catalan}. 
However, the pivot-based method requires doubled decoding time and suffers from the propagation of translation errors.

One common alternative to avoid pivoting in NMT is transfer learning~\cite{zoph2016transfer,nguyen2017transfer,kocmi2018trivial,Kim2019PivotbasedTL} which leverages a high-resource pivot$\to$target model (\emph{parent}) to initialize a low-resource source$\to$target model (\emph{child}) that is further optimized with a small amount of available parallel data.
Although this approach has achieved success in some low-resource language pairs, it still performs very poorly in extremely low-resource or zero-resource translation scenario.
Specifically, \citet{kocmi2018trivial} reports that without any child model training data, the performance of the parent model on the child test set is miserable.


In this work, we argue that the language space mismatch problem, also named \emph{domain shift problem}~\cite{fu2015transductive}, brings about the zero-shot translation failure in transfer learning. 
It is because transfer learning has no explicit training process to guarantee that the source and pivot languages share the same feature distributions, causing that the child model inherited from the parent model fails in such a situation.
For instance, as illustrated in the left of Figure \ref{fig:intro-space-map}, the points of the sentence pair with the same semantics are not overlapping in source space, resulting in that the shared decoder will generate different translations denoted by different points in target space.  
Actually, transfer learning for NMT can be viewed as a multi-domain problem where each source language forms a new domain.
Minimizing the discrepancy between the feature distributions of different source languages, i.e., different domains, will ensure the smooth transition between the parent and child models, as shown in the right of Figure \ref{fig:intro-space-map}.
One way to achieve this goal is the fine-tuning technique, which forces the model to forget the specific knowledge from parent data and learn new features from child data.
However, the \emph{domain shift problem} still exists, and the demand of parallel child data for fine-tuning heavily hinders transfer learning for NMT towards the zero-resource setting.



In this paper, we explore the transfer learning in a common zero-shot scenario where there are a lot of source$\leftrightarrow$pivot and pivot$\leftrightarrow$target parallel data but no source$\leftrightarrow$target parallel data.
In this scenario, we propose a simple but effective transfer approach, the key idea of which is to relieve the burden of the \emph{domain shift problem} by means of cross-lingual pre-training.
To this end, we firstly investigate the performance of two existing cross-lingual pre-training methods proposed by \citet{lample2019cross} in zero-shot translation scenario.
Besides, a novel pre-training method called BRidge Language Modeling (BRLM) is designed to make full use of the source$\leftrightarrow$pivot bilingual data to obtain a universal encoder for different languages.
Once the universal encoder is constructed, we only need to train the pivot$\to$target model and then test this model in source$\to$target direction directly.  
The main contributions of this paper are as follows:
\begin{itemize}
 
\item We propose a new transfer learning approach for NMT which uses the cross-lingual language model pre-training to enable a high performance on zero-shot translation.

\item We propose a novel pre-training method called BRLM, which can  effectively alleviates the distance between different source language spaces.

\item Our proposed approach significantly improves zero-shot translation performance, consistently surpassing pivoting and multilingual approaches. Meanwhile, the performance on supervised translation direction remains the same level or even better when using our method.
\end{itemize}

\section{Related Work}


In recent years, zero-shot translation in NMT has attracted widespread attention in academic research.
Existing methods are mainly divided into four categories: pivot-based method, transfer learning, multilingual NMT, and unsupervised NMT. 

\begin{itemize}

\item \textbf{Pivot-based Method} is a common strategy to obtain a source$\to$target model by introducing a pivot language.
This approach is further divided into pivoting and pivot-synthetic.
While the former firstly translates a source language into the pivot language which is later translated to the target language~\cite{kauers2002interlingua,de2006catalan,utiyama2007comparison}, the latter trains a source$\to$target model with pseudo data generated from source-pivot or pivot-target parallel data~\cite{chen2017teacher,zheng2017maximum}. 
Although the pivot-based methods can achieve not bad performance, it always falls into a computation-expensive and parameter-vast dilemma of quadratic growth in the number of source languages, and suffers from the error propagation problem~\cite{zhu2013improving}.

\item \textbf{Transfer Learning} is firstly introduced for NMT by \citet{zoph2016transfer}, which leverages a high-resource parent model to initialize the low-resource child model. 
On this basis, \citet{nguyen2017transfer} and \citet{kocmi2018trivial} use shared vocabularies for source/target language to improve transfer learning, while \citet{kim2019effective} relieve the vocabulary mismatch by mainly using cross-lingual word embedding.
Although these methods are successful in the low-resource scene, they have limited effects in zero-shot translation.

\item \textbf{Multilingual NMT} (MNMT) enables training a single model that supports translation from multiple source languages into multiple target languages, even those unseen language pairs~\cite{firat2016multi,firat2016zero,johnson2017google,al2019consistency,aharoni2019massively}.
Aside from simpler deployment, MNMT benefits from transfer learning where low-resource language pairs are trained together with high-resource ones.
However, \citet{gu2019improved} point out that MNMT for zero-shot translation easily fails, and is sensitive to the hyper-parameter setting.
Also, MNMT usually performs worse than the pivot-based method in zero-shot translation setting~\cite{arivazhagan2019missing}. 

\item \textbf{Unsupervised NMT} (UNMT) considers a harder setting, in which only large-scale monolingual corpora are available for training. 
Recently, many methods have been proposed to improve the performance of UNMT, including using denoising auto-encoder, statistic machine translation (SMT) and unsupervised pre-training ~\cite{artetxe2017unsupervised,lample2018phrase,ren2019unsupervised,lample2019cross}.
Since UNMT performs well between similar languages (e.g., English-German translation), its performance between distant languages is still far from expectation.
\end{itemize}

Our proposed method belongs to the transfer learning, but it is different from traditional transfer methods which train a parent model as starting point.
Before training a parent model, our approach fully leverages cross-lingual pre-training methods to make all source languages share the same feature space and thus enables a smooth transition for zero-shot translation.

\begin{figure*}[htbp]
\centering
\includegraphics[width=0.96\textwidth]{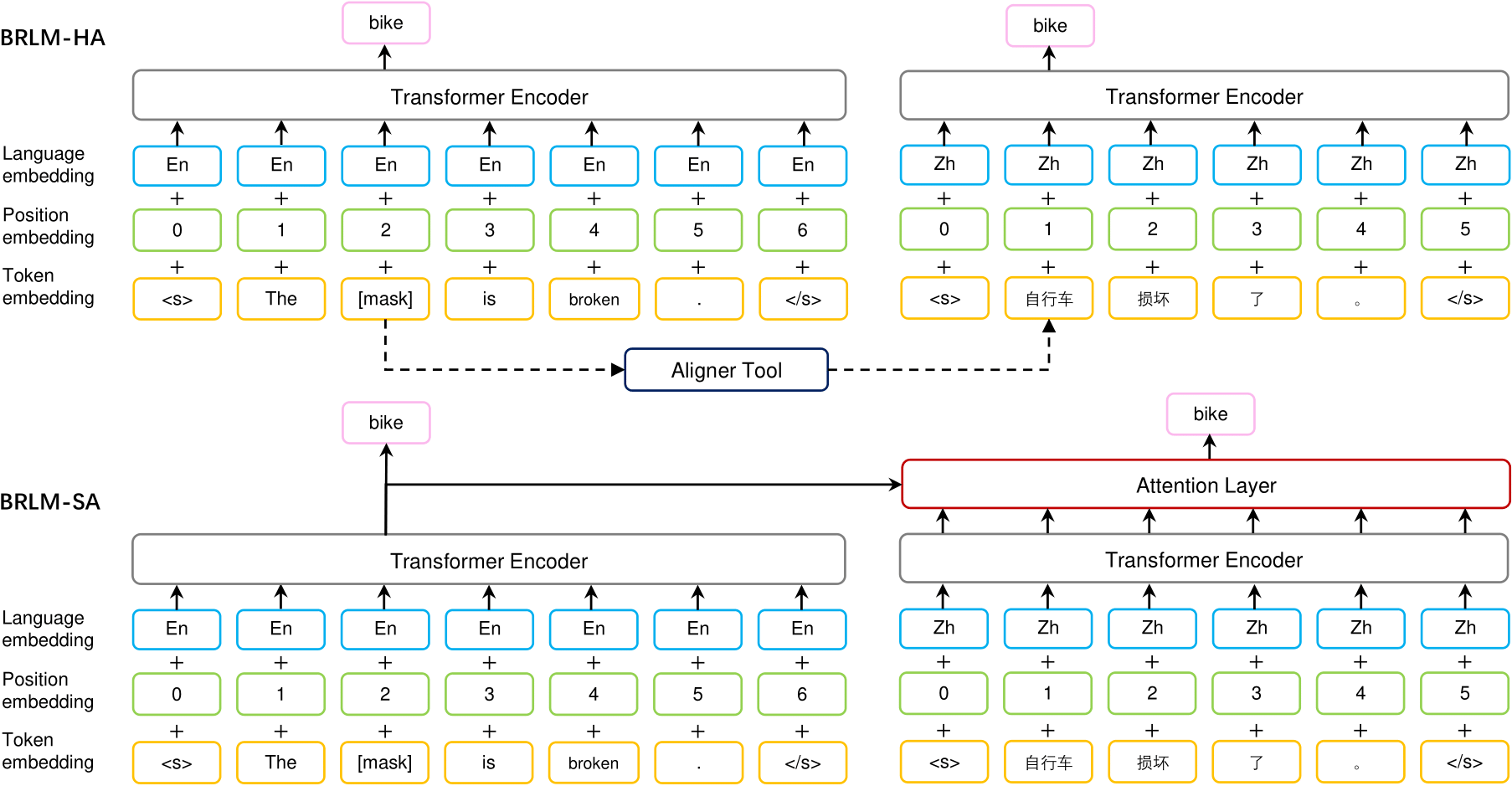}
\caption{The overview of BRidge Language Modeling (BRLM). The BRLM extends MLM~\cite{lample2019cross} to pairs of parallel sentences and leverages explicit alignment information obtained by external aligner tool or additional attention layer to encourage word representation alignment across different languages.}
\label{fig:model-framework}
\end{figure*}

\section{Approach}

In this section, we will present a cross-lingual pre-training based transfer approach.
This method is designed for a common zero-shot scenario where there are a lot of source$\leftrightarrow$pivot and pivot$\leftrightarrow$target bilingual data but no source$\leftrightarrow$target parallel data, and the whole training process can be summarized as follows step by step:
\begin{itemize}
\item Pre-train a universal encoder with source/pivot monolingual or source$\leftrightarrow$pivot bilingual data.
\item Train a pivot$\to$target parent model built on the pre-trained universal encoder with the available parallel data. During the training process, we freeze several layers of the pre-trained universal encoder to avoid the degeneracy issue~\cite{Howard2018UniversalLM}.
\item Directly translate source sentences into target sentences with the parent model, which benefits from the availability of the universal encoder. 
\end{itemize}
The key difficulty of this method is to ensure the intermediate representations of the universal encoder are language invariant.
In the rest of this section, we first present two existing methods yet to be explored in zero-shot translation, and then propose a straightforward but effective cross-lingual pre-training method.
In the end, we present the whole training and inference protocol for transfer. 


\subsection{Masked and Translation Language Model Pretraining}

Two existing cross-lingual pre-training methods, Masked Language Modeling (MLM) and Translation Language Modeling (TLM), have shown their effectiveness on XNLI cross-lingual classification task~\cite{lample2019cross,Huang2019UnicoderAU}, but these methods have not been well studied on cross-lingual generation tasks in zero-shot condition.
We attempt to take advantage of the cross-lingual ability of the two methods for zero-shot translation.

Specifically, MLM adopts the Cloze objective of BERT~\cite{devlin2018bert} and predicts the masked words that are randomly selected and replaced with [MASK] token on monolingual corpus. 
In practice, MLM takes different language monolingual corpora as input to find features shared across different languages.
With this method, word pieces shared in all languages have been mapped into a shared space, which makes the sentence representations across different languages close~\cite{DBLP:journals/corr/abs-1906-01502}.

Since MLM objective is unsupervised and only requires monolingual data, TLM is designed to leverage parallel data when it is available.
Actually, TLM is a simple extension of MLM, with the difference that TLM concatenates sentence pair into a whole sentence, and then randomly masks words in both the source and target sentences.
In this way, the model can either attend to surrounding words or to the translation sentence, implicitly encouraging the model to align the source and target language representations.
Note that although each sentence pair is formed into one sentence, the positions of the target sentence are reset to count form zero.

\subsection{Bridge Language Model Pretraining}

Aside from MLM and TLM, we propose BRidge Language Modeling (BRLM) to further obtain word-level representation alignment between different languages. 
This method is inspired by the assumption that if the feature spaces of different languages are aligned very well, the masked words in the corrupted sentence can also be guessed by the context of the correspondingly aligned words on the other side. 
To achieve this goal, BRLM is designed to strengthen the ability to infer words across languages based on alignment information, instead of inferring words within monolingual sentence as in MLM or within the pseudo sentence formed by concatenating sentence pair as in TLM.

As illustrated in Figure \ref{fig:model-framework}, BRLM stacks shared encoder over both side sentences separately. In particular, we design two network structures for BRLM, which are divided into Hard Alignment (BRLM-HA) and Soft Alignment (BRLM-SA) according to the way of generating the alignment information.
These two structures actually extend MLM into a bilingual scenario, with the difference that BRLM leverages external aligner tool or additional attention layer to explicitly introduce alignment information during model training.

\begin{itemize}

\item \textbf{Hard Alignment (BRLM-HA).}  We first use external aligner tool on source$\leftrightarrow$pivot parallel data to extract the alignment information of sentence pair.
During model training, given source$\leftrightarrow$pivot sentence pair, BRLM-HA randomly masks some words in source sentence and leverages alignment information to obtain the aligned words in pivot sentence for masked words.
Based on the processed input, BRLM-HA adopts the Transformer~\cite{vaswani2017attention} encoder to gain the hidden states for source and pivot sentences respectively.
Then the training objective of BRLM-HA is to predict the masked words by not only the surrounding words in source sentence but also the encoder outputs of the aligned words. 
Note that this training process is also carried out in a symmetric situation, in which we mask some words in pivot sentence and obtain the aligned words in the source sentence.

\item \textbf{Soft Alignment (BRLM-SA).} Instead of using external aligner tool, BRLM-SA introduces an additional attention layer to learn the alignment information together with model training.
In this way, BRLM-SA avoids the effect caused by external wrong alignment information and enables many-to-one soft alignment during model training.
Similar with BRLM-HA, the training objective of BRLM-SA is to predict the masked words by not only the surrounding words in source sentence but also the outputs of attention layer.
In our implementation, the attention layer is a multi-head attention layer adopted in Transformer, where the queries come from the masked source sentence, the keys and values come from the pivot sentence.

\end{itemize} 

\noindent  In principle, MLM and TLM can learn some implicit alignment information during model training.
However, the alignment process in MLM is inefficient since the shared word pieces only account for a small proportion of the whole corpus, resulting in the difficulty of expanding the shared information to align the whole corpus.
TLM also lacks effort in alignment between the source and target sentences since TLM concatenates the sentence pair into one sequence, making the explicit alignment between the source and target infeasible.
BRLM fully utilizes the alignment information to obtain better word-level representation alignment between different languages, which better relieves the burden of the \emph{domain shift problem}.

\subsection{Transfer Protocol}

We consider the typical zero-shot translation scenario in which a high resource pivot language has parallel data with both source and target languages, while source and target languages has no parallel data between themselves. 
Our proposed cross-lingual pretraining based transfer approach for source$\rightarrow$target zero-shot translation is mainly divided into two phrases: the pretraining phase and the transfer phase.

In the pretraining phase, we first pretrain MLM on monolingual corpora of both source and pivot languages, and continue to pretrain TLM or the proposed BRLM on the available parallel data between source and pivot languages, in order to build a cross-lingual encoder shared by the source and pivot languages.

In the transfer phase, we train pivot$\rightarrow$target NMT model initialized by the cross-lingually pre-trained encoder, and finally transfer the trained NMT model to source$\rightarrow$target translation thanks to the shared encoder. Note that during training pivot$\rightarrow$target NMT model, we freeze several layers of the cross-lingually pre-trained encoder to avoid the degeneracy issue.

For the more complicated scenario that either the source side or the target side has multiple languages, the encoder and the decoder are also shared across each side languages for efficient deployment of translation between multiple languages.

\section{Experiments}

\begin{table}[t]
\centering
\small
\begin{tabular}{c|c|c|c|c}
\bottomrule[1.2pt]
Corpus & Language & Train & Dev  & Test  \\
\hline
\multirow{3}{*}{Europarl} & De-En,En-Fr & 1M,1M &2,000 & 2,000\\
                          & Fr-En,En-Es & 1M,1M &2,000 & 2,000\\
                          & Ro-En,En-De & 0.6M,1.5M & 2,000 & 1,000\\
\hline        
\multirow{1}{*}{MultiUN}& \makecell{Ar-En,En-Es\\En-Ru} & \makecell{9.7M,11.3M \\11.6M}& 4,000 & 4,000 \\
\toprule[1.2pt]
\end{tabular}
\caption{Data Statistics.}
\label{tab-data}
\end{table}

\begin{table*}[t]
\centering
\begin{tabular}{c|c|c|c|c|c|c}
\bottomrule[1.2pt]
Europarl & \multicolumn{2}{c|}{Fr $\to$ En $\to$ Es} & \multicolumn{2}{c|}{De $\to$ En $\to$ Fr} & \multicolumn{2}{c}{Ro $\to$ En $\to$ De} \\ \cline{1-7} 
Direction &  Fr $\to$ Es   & En $\to$ Es &    De $\to$ Fr  & En $\to$ Fr & Ro $\to$ De &   En $\to$ De         \\ \hline
\multicolumn{7}{c}{Baselines} \\ \hline
Cross-lingual Transfer~\cite{kim2019effective} &18.45 & 34.01 & 9.86 & 34.05 &2.02& 23.61 \\
MNMT\cite{johnson2017google}  &27.12&34.69&21.36&33.87&9.31&24.09 \\
MNMT$_{\rm Agreement}$~\cite{al2019consistency} & 29.91 &33.80&24.45 &32.55 & - & - \\ 
Pivoting &            32.25&34.01&27.79& 34.05&14.74 & 23.61 \\\hline
\multicolumn{7}{c}{Proposed Cross-lingual Pretraining Based Transfer} \\ \hline
MLM   &              35.96&34.83&27.61&35.66&12.64&22.04               \\
MLM+TLM &              36.78&34.73&29.45&35.33&14.39&24.96              \\  
MLM+BRLM-HA&           36.30&34.98&29.91&34.99&14.21&24.26               \\ 
MLM+BRLM-SA&           \textbf{37.02}&34.92&\textbf{30.66}&35.91&\textbf{15.62}&24.95               \\ 
\toprule[1.2pt]
\end{tabular}
\caption{Results on Europarl test sets. Three pivot settings are conducted in our experiments. In each setting, the left column presents the zero-shot performances (source$\rightarrow$target), and the right column denotes the performances in the supervised parent model direction (pivot$\rightarrow$target). }
\label{tab-euro-results}  
\end{table*}

\subsection{Setup}

We evaluate our cross-lingual pre-training based transfer approach against several strong baselines on two public datatsets, Europarl~\cite{koehn2005europarl} and MultiUN~\cite{eisele2010multiun}, which contain multi-parallel evaluation data to assess the zero-shot performance.
In all experiments, we use BLEU as the automatic metric for translation evaluation.\footnote{We calculate BLEU scores with the \emph{multi-bleu.perl} script.}

\paragraph{Datasets.} The statistics of Europarl and MultiUN corpora are summarized in Table~\ref{tab-data}. 
For Europarl corpus, we evaluate on French-English-Spanish (Fr-En-Es), German-English-French (De-En-Fr) and Romanian-English-German (Ro-En-De), where English acts as the pivot language, its left side is the source language, and its right side is the target language. 
We remove the multi-parallel sentences between different training corpora to ensure zero-shot settings. 
We use the devtest2006 as the validation set and the test2006 as the test set for Fr$\to$Es and De$\to$Fr. 
For distant language pair Ro$\to$De, we extract 1,000 overlapping sentences from newstest2016 as the test set and the 2,000 overlapping sentences split from the training set as the validation set since there is no official validation and test sets.
For vocabulary, we use 60K sub-word tokens based on Byte Pair Encoding (BPE) ~\cite{sennrich2015neural}. 

For MultiUN corpus, we use four languages: English (En) is set as the pivot language, which has parallel data with other three languages which do not have parallel data between each other. 
The three languages are Arabic (Ar), Spanish (Es), and Russian (Ru), and mutual translation between themselves constitutes six zero-shot translation direction for evaluation. 
We use 80K BPE splits as the vocabulary.
Note that all sentences are tokenized by the \emph{tokenize.perl}\footnote{https://github.com/moses-smt/mosesdecoder/blob/RELEASE-3.0/scripts/tokenizer/tokenizer.perl} script, and we lowercase all data to avoid a large vocabulary for the MultiUN corpus. 

\paragraph{Experimental Details.}
We use traditional transfer learning, pivot-based method and multilingual NMT as our baselines.
For the fair comparison, the Transformer-big model with 1024 embedding/hidden units, 4096 feed-forward filter size, 6 layers and 8 heads per layer is adopted for all translation models in our experiments. 
We set the batch size to 2400 per batch and limit sentence length to 100 BPE tokens. 
We set the $\text{attn}\_\text{drop}=0$ (a dropout rate on each attention head), which is favorable to the zero-shot translation and has no effect on supervised translation directions~\cite{gu2019improved}.
For the model initialization, we use Facebook's cross-lingual pretrained models released by XLM\footnote{https://github.com/facebookresearch/XLM} to initialize the encoder part, and the rest parameters are initialized with xavier uniform. 
We employ the Adam optimizer with $\text{lr}=0.0001$, $t_{\text{warm}\_\text{up}}=4000$ and $\text{dropout}=0.1$.
At decoding time, we generate greedily with length penalty $\alpha=1.0$.

Regarding MLM, TLM and BRLM, as mentioned in the pre-training phase of transfer protocol, we first pre-train MLM on monolingual data of both source and pivot languages, then leverage the parameters of MLM to initialize TLM and the proposed BRLM, which are continued to be optimized with source-pivot bilingual data.
In our experiments, we use MLM+TLM, MLM+BRLM to represent this training process.
For the masking strategy during training, following \citet{devlin2018bert}, $15\%$ of BPE tokens are selected to be masked. Among the selected tokens, $80\%$ of them are replaced with [MASK] token, $10\%$ are replaced with a random BPE token, and $10\%$ unchanged.
The prediction accuracy of masked words is used as a stopping criterion in the pre-training stage.
Besides, we use \emph{fastalign} tool~\cite{dyer2013simple} to extract word alignments for BRLM-HA. 




\begin{table*}[t]
\centering
\begin{tabular}{c|c|c|c|c|c|c|c|c}
\bottomrule[1.2pt]
MultiUN & \multicolumn{8}{c}{ Ar,Es,Ru $\leftrightarrow$ En } \\ \cline{1-9} 
Direction   &  Ar $\to$ Es & Es $\to$ Ar & Ar $\to$ Ru &  Ru $\to$ Ar & Es $\to$ Ru  & Ru $\to$ Es  &  A-ZST & A-ST\\ \hline
\multicolumn{9}{c}{ Baselines } \\ \cline{1-9}
Cross-lingual Transfer & 10.26 & 12.44 & 4.58 & 4.42 & 13.80 & 7.93 & 8.90 & 44.73  \\ 
MNMT\cite{johnson2017google}  & 27.40 & 20.18 & 15.12 &16.19 &17.88&27.93&20.78&43.95 \\ 
Pivoting$_{\rm m}$ &42.29&30.15&27.23& 26.16 &29.57&40.08&32.58&43.95\\  \hline
\multicolumn{9}{c}{ Proposed Cross-lingual Pretraining Based Transfer } \\ \cline{1-9} 
MLM   &16.50&23.41&9.61&14.23 &22.80&23.66&18.36&44.25\\ 
MLM+TLM    &25.98&26.55&16.84&20.07 &25.91&29.52&24.14&43.71 \\ 
MLM+BRLM-HA  &29.05&27.58&18.10&20.42 &25.39&30.96&25.25& 44.67\\
MLM+BRLM-SA  &36.01&31.08&25.49&25.06 &30.47&36.01&30.68& 44.54 \\ \hline
\multicolumn{9}{c}{ Adding Back Translation } \\ \cline{1-9}
MNMT*~\cite{gu2019improved} &39.72&28.05&24.67&24.43 &27.41&38.01&30.38&43.98 \\ \hline
MLM  &40.98&31.53&26.06&26.69 &31.28&40.02&32.76&44.28 \\
MLM+TLM &41.15&29.77&27.61&27.74 &31.02&40.37&32.39&44.14 \\
MLM+BRLM-HA &41.74&31.89&27.24&27.54&31.29&40.34&33.35&44.52 \\
MLM+BRLM-SA &\textbf{44.17}&\textbf{33.20}&\textbf{29.01}&\textbf{28.91}&\textbf{32.53}&\textbf{41.93}&\textbf{34.95}&45.49 \\ 
\toprule[1.2pt]
\end{tabular}
\caption{Results on MultiUN test sets. The six zero-shot translation directions are evaluated. The column ``A-ZST" reports averaged BLEU of zero-shot translation, while the column ``A-ST" reports averaged BLEU of supervised pivot$\rightarrow$target direction.}
\label{tab-un-results}  
\end{table*}

\begin{figure*}[t]
\centering
\subfigure[Fr-En]{
        \includegraphics[width=0.30\linewidth]{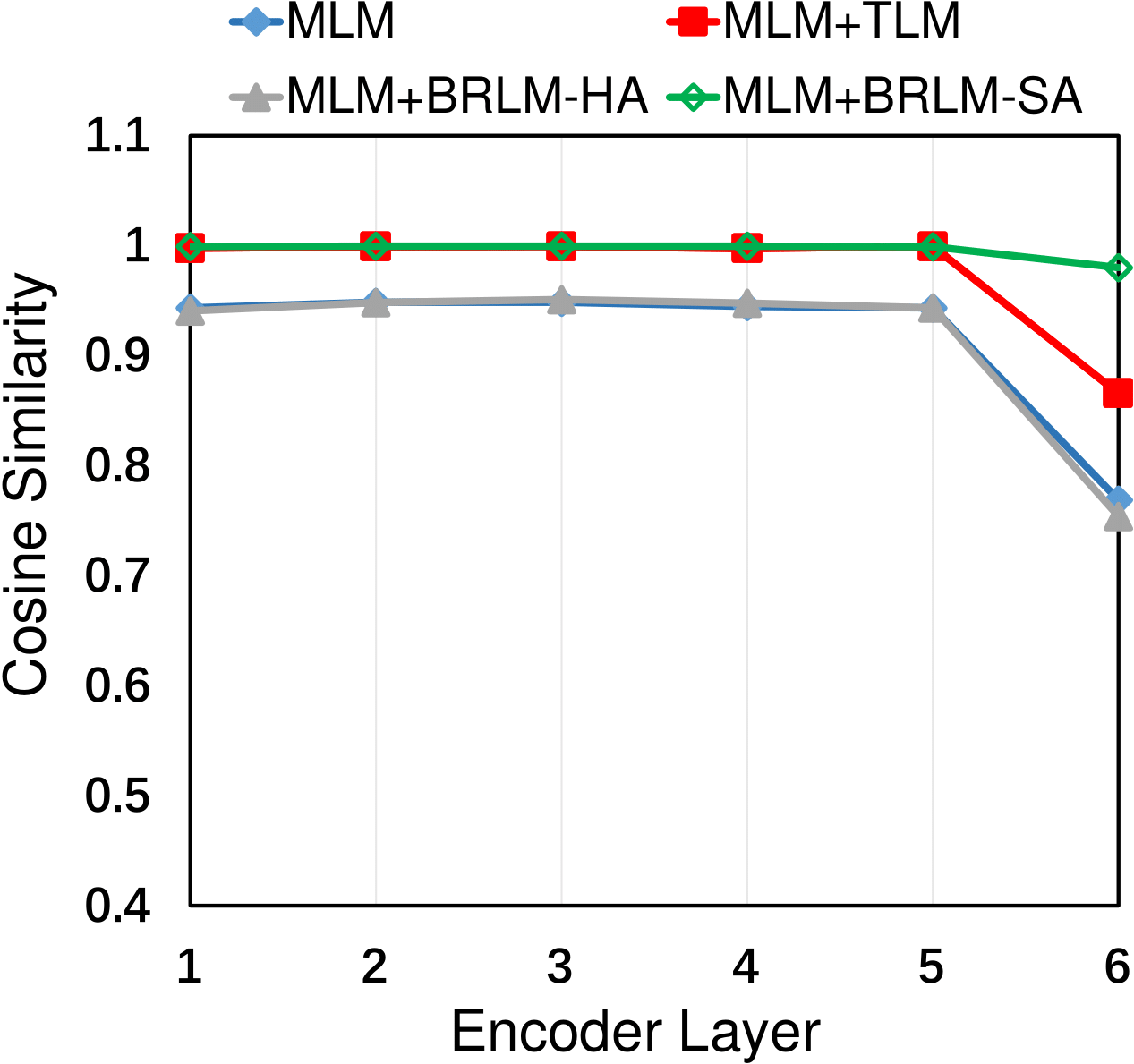}
}
\hspace{1mm}
\subfigure[De-En]{
        \includegraphics[width=0.30\linewidth]{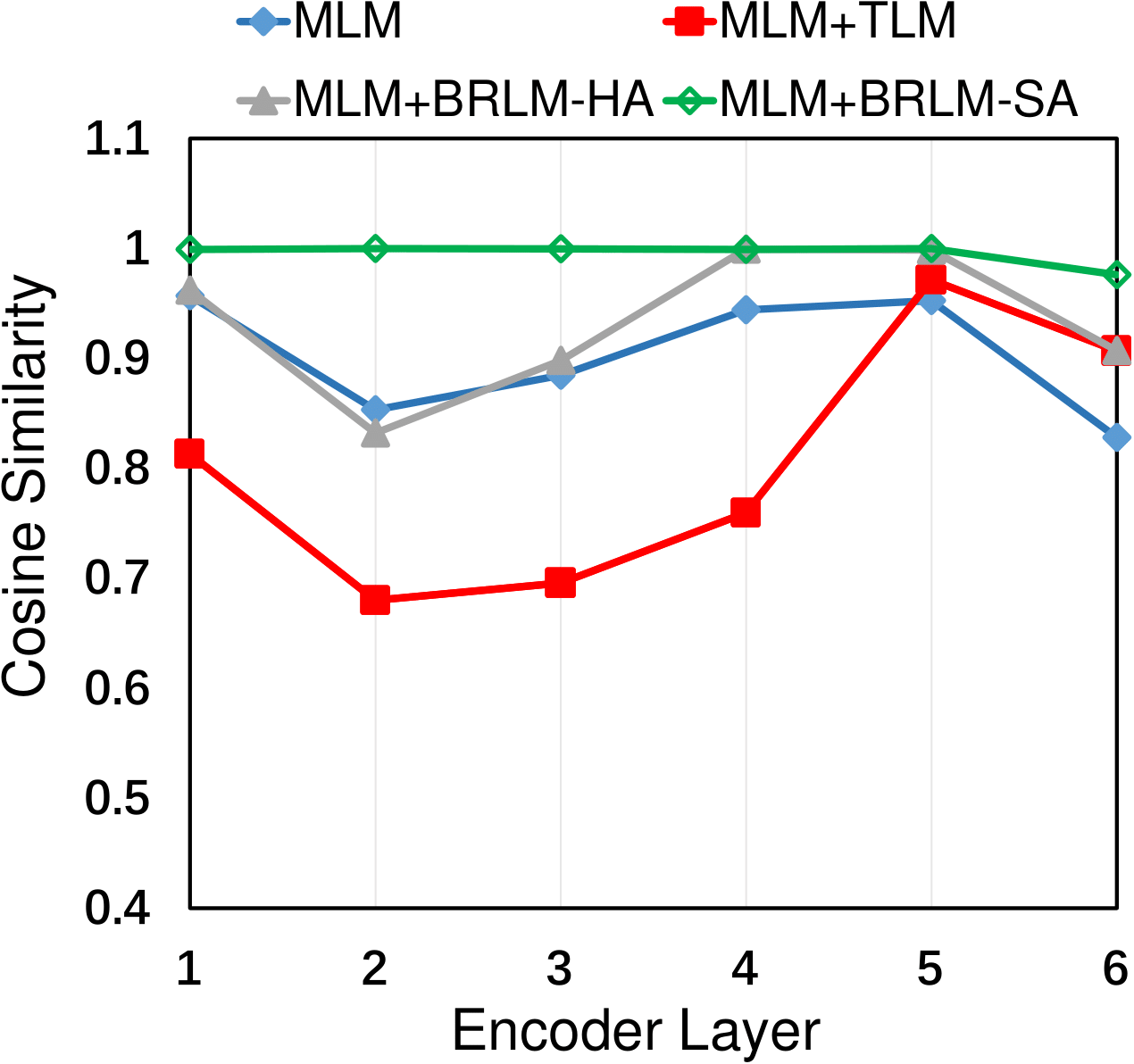}
}
\hspace{1mm}
\subfigure[Ro-En]{
        \includegraphics[width=0.30\linewidth]{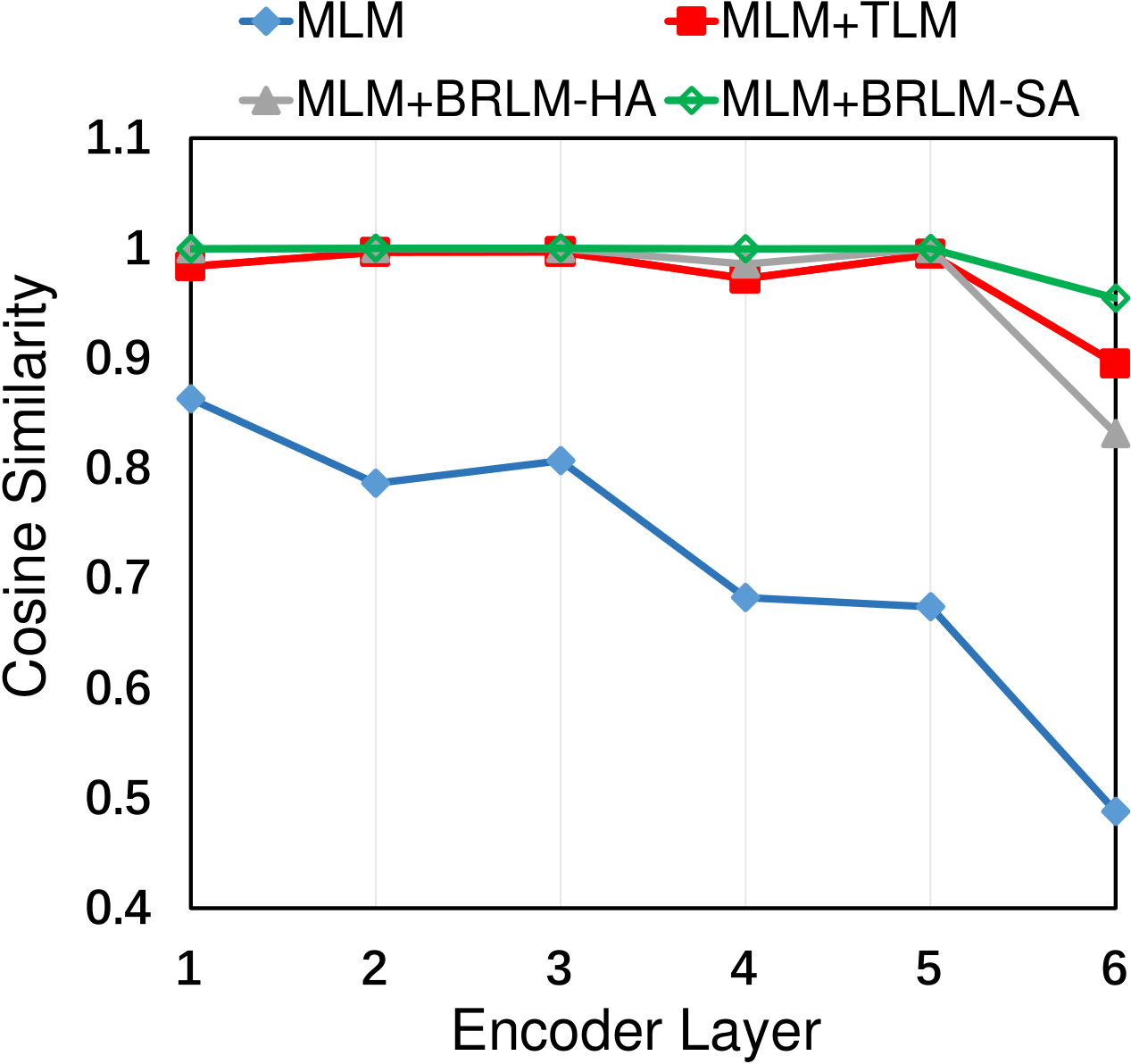}
}
\caption{Cosine similarity between sentence representation of each encoder layer across all source-pivot sentence pairs in the Europarl validation set. }
\label{fig:cosine-similarity}
\end{figure*}

\subsection{Main Results}

Table \ref{tab-euro-results} and \ref{tab-un-results} report zero-shot results on Europarl and Multi-UN evaluation sets, respectively. We compare our approaches with related approaches of pivoting, multilingual NMT (MNMT)~\cite{johnson2017google}, and cross-lingual transfer without pretraining~\cite{kim2019effective}. The results show that our approaches consistently outperform other approaches across languages and datasets, especially surpass pivoting, which is a strong baseline in the zero-shot scenario that multilingual NMT systems often fail to beat~ \cite{johnson2017google,al2019consistency,arivazhagan2019missing}. Pivoting translates source to pivot then to target in two steps, causing inefficient translation process. Our approaches use one encoder-decoder model to translate between any zero-shot directions, which is more efficient than pivoting. Regarding the comparison between transfer approaches, our cross-lingual pretraining based transfer outperforms transfer method that does not use pretraining by a large margin. 


\paragraph{Results on Europarl Dataset.} Regarding comparison between the baselines in table \ref{tab-euro-results}, we find that pivoting is the strongest baseline that has significant advantage over other two baselines. Cross-lingual transfer for languages without shared vocabularies \cite{kim2019effective} manifests the worst performance because of not using source$\leftrightarrow$pivot parallel data, which is utilized as beneficial supervised signal for the other two baselines.

Our best approach of MLM+BRLM-SA achieves the significant superior performance to all baselines in the zero-shot directions, improving by 0.9-4.8 BLEU points over the strong pivoting. 
Meanwhile, in the supervised direction of pivot$\rightarrow$target, our approaches performs even better than the original supervised Transformer thanks to the shared encoder trained on both large-scale monolingual data and parallel data between multiple languages. 

MLM alone that does not use source$\leftrightarrow$pivot parallel data performs much better than the cross-lingual transfer, and achieves comparable results to pivoting. 
When MLM is combined with TLM or the proposed BRLM, the performance is further improved. MLM+BRLM-SA performs the best, and is better than MLM+BRLM-HA indicating that soft alignment is helpful than hard alignment for the cross-lingual pretraining.

\paragraph{Results on MultiUN Dataset.} Like experimental results on Europarl, MLM+BRLM-SA performs the best among all proposed cross-lingual pretraining based transfer approaches as shown in Table \ref{tab-un-results}. When comparing systems consisting of one encoder-decoder model for all zero-shot translation, our approaches performs significantly better than MNMT \cite{johnson2017google}.

Although it is challenging for one model to translate all zero-shot directions between multiple distant language pairs of MultiUN, MLM+BRLM-SA still achieves better performances on Es $\rightarrow$ Ar and Es $\rightarrow$ Ru than strong pivoting$_{\rm m}$, which uses MNMT to translate source to pivot then to target in two separate steps with each step receiving supervised signal of parallel corpora. Our approaches surpass pivoting$_{\rm m}$ in all zero-shot directions by adding back translation \cite{sennrich2015neural} to generate pseudo parallel sentences for all zero-shot directions based on our pretrained models such as MLM+BRLM-SA, and further training our universal encoder-decoder model with these pseudo data. \citeauthor{gu2019improved} \shortcite{gu2019improved} introduces back translation into MNMT, while we adopt it in our transfer approaches. Finally, our best MLM+BRLM-SA with back translation outperforms pivoting$_{\rm m}$ by 2.4 BLEU points averagely, and outperforms MNMT \cite{gu2019improved} by 4.6 BLEU points averagely. Again, in supervised translation directions, MLM+BRLM-SA with back translation also achieves better performance than the original supervised Transformer.

\subsection{Analysis}


\paragraph{Sentence Representation.} We first evaluate the representational invariance across languages for all cross-lingual pre-training methods. 
Following \citet{arivazhagan2019missing}, we adopt max-pooling operation to collect the sentence representation of each encoder layer for all source-pivot sentence pairs in the Europarl validation sets.
Then we calculate the cosine similarity for each sentence pair and average all cosine scores.
As shown in Figure \ref{fig:cosine-similarity}, we can observe that, MLM+BRLM-SA has the most stable and similar cross-lingual representations of sentence pairs on all layers, while it achieves the best performance in zero-shot translation. 
This demonstrates that better cross-lingual representations can benefit for the process of transfer learning.
Besides, MLM+BRLM-HA is not as superior as MLM+BRLM-SA and even worse than MLM+TLM on Fr-En, since MLM+BRLM-HA may suffer from the wrong alignment knowledge from an external aligner tool.
We also find an interesting phenomenon that as the number of layers increases, the cosine similarity decreases.

\begin{figure}[t]
\centering
\subfigure[MLM]{
        \includegraphics[width=0.40\linewidth,height=32mm]{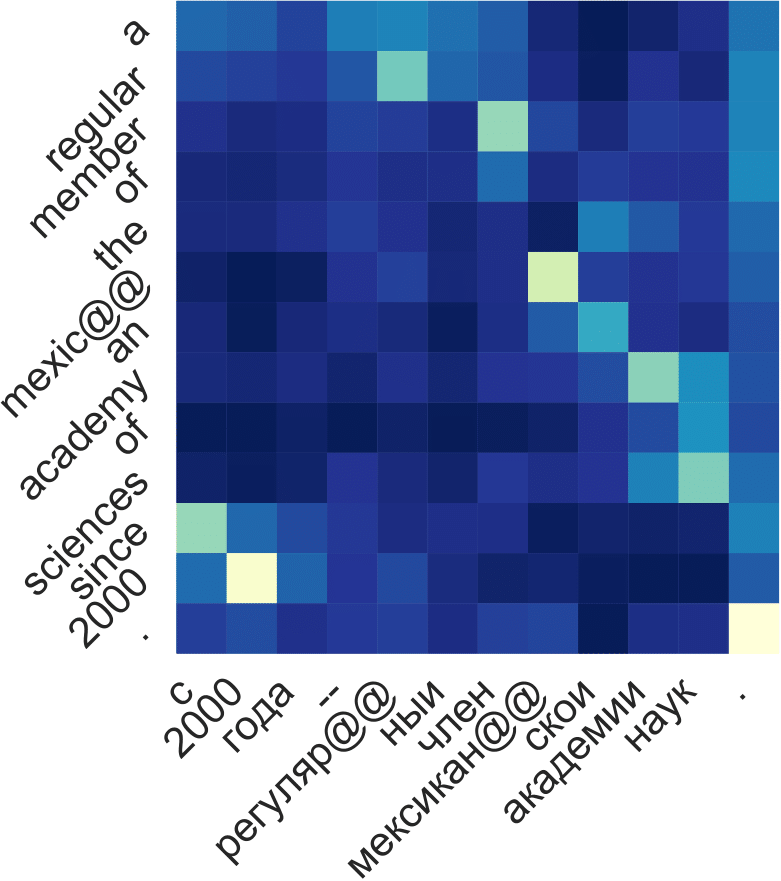}
}
\hspace{0in}
\subfigure[MLM+TLM]{
        \includegraphics[width=0.40\linewidth,height=32mm]{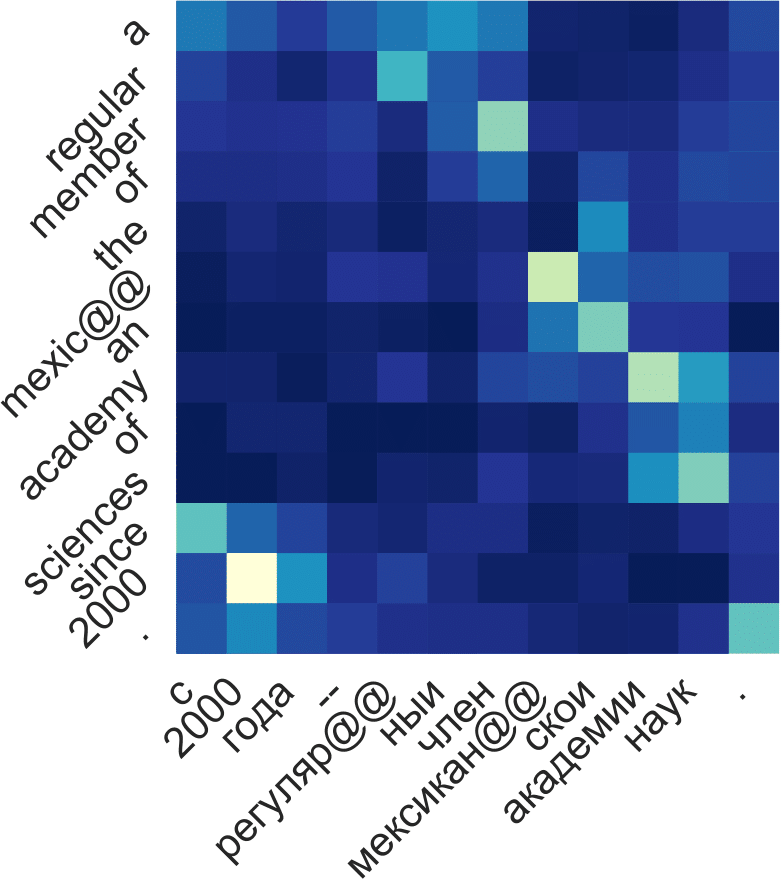}
}

\subfigure[MLM+BRLM-HA]{
    \includegraphics[width=0.40\linewidth,height=32mm]{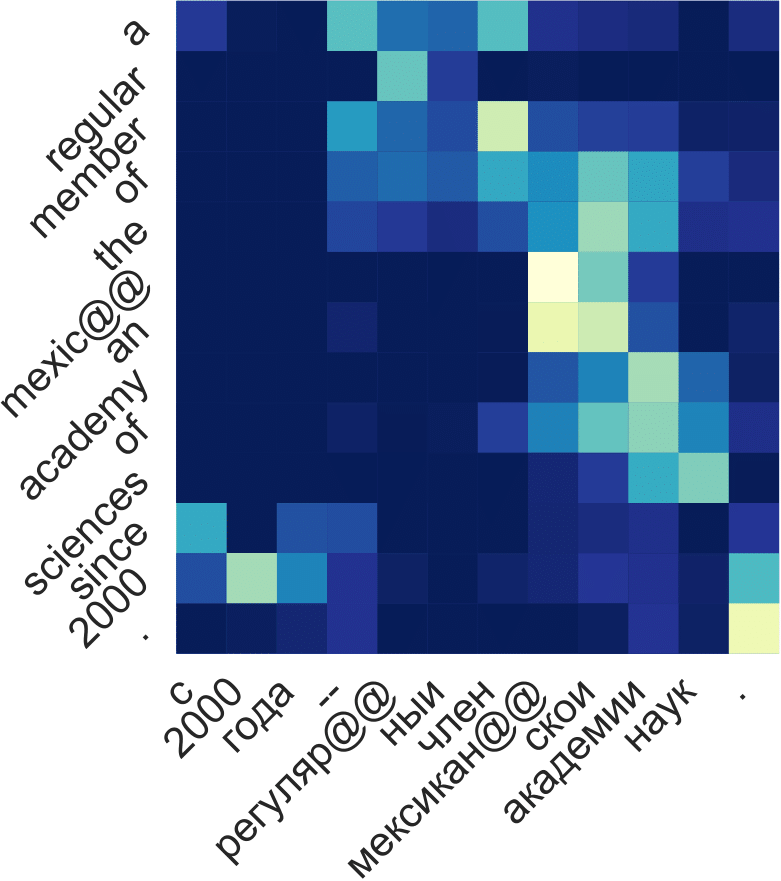}
}
\hspace{0in}
\subfigure[MLM+BRLM-SA]{
    \includegraphics[width=0.40\linewidth,height=32mm]{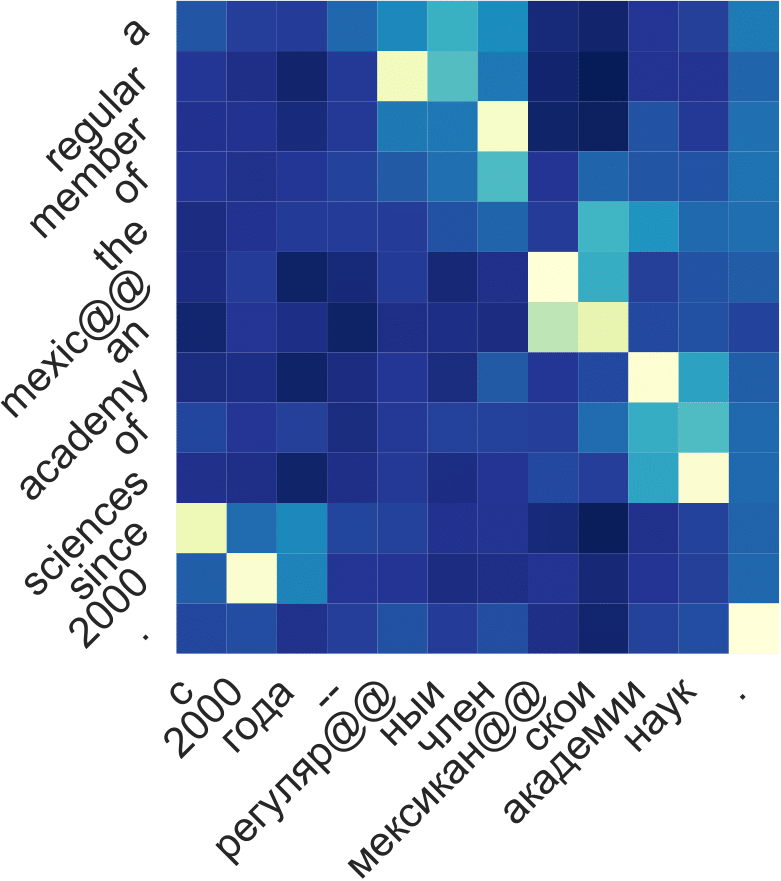}
}
\caption{Cosine similarity visualization at word level given an English-Russian sentence pair from the MultiUN validation sets. Brighter indicates higher similarity.}
\label{fig:word-level-diff}
\end{figure}

\paragraph{Contextualized Word Representation.}  
We further sample an English-Russian sentence pair from the MultiUN validation sets and visualize the cosine similarity between hidden states of the top encoder layer to further investigate the difference of all cross-lingual pre-training methods. 
As shown in Figure \ref{fig:word-level-diff}, the hidden states generated by MLM+BRLM-SA have higher similarity for two aligned words.
It indicates that MLM+BRLM-SA can gain better word-level representation alignment between source and pivot languages, which better relieves the burden of the \emph{domain shift problem}.   

\paragraph{The Effect of Freezing Parameters.} To freeze parameters is a common strategy to avoid catastrophic forgetting in transfer learning~\cite{Howard2018UniversalLM}.
Table \ref{Freeze Parameters} shows the performance of transfer learning with freezing different layers on MultiUN test set, in which En$\to$Ru denotes the parent model, Ar$\to$Ru and Es$\to$Ru are two child models, and all models are based on MLM+BRLM-SA.
We can find that updating all parameters during training will cause a notable drop on the zero-shot direction due to the catastrophic forgetting.
On the contrary, freezing all the parameters leads to the decline on supervised direction because the language features extracted during pre-training is not sufficient for MT task.
Freezing the first four layers of the transformer shows the best performance and keeps the balance between pre-training and fine-tuning. 

\begin{table}[t]
\centering
\begin{tabular}{c|c|c|c}
\bottomrule[1.2pt]
Freezing Layers & En $\to$ Ru & Ar $\to$ Ru & Es $\to$ Ru \\ \hline
None & 37.80 & 16.09& 19.80 \\
2 & 37.79 & 21.47 &  28.35\\
4 & 37.55 & \textbf{25.49} &  \textbf{30.47}\\
6 & 35.31 & 22.90 & 28.22 \\ 
\toprule[1.2pt]
\end{tabular}
\caption{BLEU score of freezing different layers. The number in Freezing Layers column denotes that the number   of encoder layers  will not be updated. }
\label{Freeze Parameters} 
\end{table}



\section{Conclusion}
In this paper, we propose a cross-lingual pretraining based transfer approach for the challenging zero-shot translation task, in which source and target languages have no parallel data, while they both have parallel data with a high resource pivot language. With the aim of building the language invariant representation between source and pivot languages for smooth transfer of the parent model of pivot$\rightarrow$target direction to the child model of source$\rightarrow$target direction, we introduce one monolingual pretraining method and two bilingual pretraining methods to construct an universal encoder for the source and pivot languages. Experiments on public datasets show that our approaches significantly outperforms several strong baseline systems, and manifest the language invariance characteristics in both sentence level and word level neural representations.

\section{Acknowledgments}
We would like to thank the anonymous reviewers for the helpful comments. 
This work was supported by National Key R\&D Program of China (Grant No. 2016YFE0132100), National Natural Science Foundation of China (Grant No. 61525205, 61673289).
This work was also partially supported by Alibaba Group through Alibaba Innovative Research Program and  the Priority Academic Program Development (PAPD) of Jiangsu Higher Education Institutions.


\bibliographystyle{aaai}
\bibliography{aaai20}

\begin{thebibliography}{}

\bibitem[\protect\citeauthoryear{Aharoni, Johnson, and
  Firat}{2019}]{aharoni2019massively}
Aharoni, R.; Johnson, M.; and Firat, O.
\newblock 2019.
\newblock Massively multilingual neural machine translation.
\newblock In {\em NAACL-HLT}.

\bibitem[\protect\citeauthoryear{Al-Shedivat and
  Parikh}{2019}]{al2019consistency}
Al-Shedivat, M., and Parikh, A.~P.
\newblock 2019.
\newblock Consistency by agreement in zero-shot neural machine translation.
\newblock In {\em NAACL-HLT}.

\bibitem[\protect\citeauthoryear{Arivazhagan \bgroup et al\mbox.\egroup
  }{2018}]{arivazhagan2019missing}
Arivazhagan, N.; Bapna, A.; Firat, O.; Aharoni, R.; Johnson, M.; and Macherey,
  W.
\newblock 2018.
\newblock The missing ingredient in zero-shot neural machine translation.
\newblock {\em ArXiv} abs/1903.07091.

\bibitem[\protect\citeauthoryear{Artetxe \bgroup et al\mbox.\egroup
  }{2017}]{artetxe2017unsupervised}
Artetxe, M.; Labaka, G.; Agirre, E.; and Cho, K.
\newblock 2017.
\newblock Unsupervised neural machine translation.
\newblock {\em ArXiv} abs/1710.11041.

\bibitem[\protect\citeauthoryear{Chen \bgroup et al\mbox.\egroup
  }{2017}]{chen2017teacher}
Chen, Y.; Liu, Y.~P.; Cheng, Y.; and Li, V. O.~K.
\newblock 2017.
\newblock A teacher-student framework for zero-resource neural machine
  translation.
\newblock In {\em ACL}.

\bibitem[\protect\citeauthoryear{de Gispert and
  Mari{\~n}o}{2006}]{de2006catalan}
de~Gispert, A., and Mari{\~n}o, J.~B.
\newblock 2006.
\newblock Catalan-english statistical machine translation without parallel
  corpus : Bridging through spanish.

\bibitem[\protect\citeauthoryear{Devlin \bgroup et al\mbox.\egroup
  }{2018}]{devlin2018bert}
Devlin, J.; Chang, M.-W.; Lee, K.; and Toutanova, K.
\newblock 2018.
\newblock Bert: Pre-training of deep bidirectional transformers for language
  understanding.
\newblock In {\em NAACL-HLT}.

\bibitem[\protect\citeauthoryear{Dyer, Chahuneau, and
  Smith}{2013}]{dyer2013simple}
Dyer, C.; Chahuneau, V.; and Smith, N.~A.
\newblock 2013.
\newblock A simple, fast, and effective reparameterization of ibm model 2.
\newblock In {\em HLT-NAACL}.

\bibitem[\protect\citeauthoryear{Eisele and Chen}{2010}]{eisele2010multiun}
Eisele, A., and Chen, Y.
\newblock 2010.
\newblock Multiun: A multilingual corpus from united nation documents.
\newblock In {\em LREC}.

\bibitem[\protect\citeauthoryear{Firat \bgroup et al\mbox.\egroup
  }{2016}]{firat2016zero}
Firat, O.; Sankaran, B.; Al-Onaizan, Y.; Yarman-Vural, F.~T.; and Cho, K.
\newblock 2016.
\newblock Zero-resource translation with multi-lingual neural machine
  translation.
\newblock In {\em EMNLP}.

\bibitem[\protect\citeauthoryear{Firat, Cho, and Bengio}{2016}]{firat2016multi}
Firat, O.; Cho, K.; and Bengio, Y.
\newblock 2016.
\newblock Multi-way, multilingual neural machine translation with a shared
  attention mechanism.
\newblock In {\em HLT-NAACL}.

\bibitem[\protect\citeauthoryear{Fu \bgroup et al\mbox.\egroup
  }{2015}]{fu2015transductive}
Fu, Y.; Hospedales, T.~M.; Xiang, T.~Y.; and Gong, S.
\newblock 2015.
\newblock Transductive multi-view zero-shot learning.
\newblock {\em IEEE Transactions on Pattern Analysis and Machine Intelligence}
  37:2332--2345.

\bibitem[\protect\citeauthoryear{Gu \bgroup et al\mbox.\egroup
  }{2019}]{gu2019improved}
Gu, J.; Wang, Y.; Cho, K.; and Li, V. O.~K.
\newblock 2019.
\newblock Improved zero-shot neural machine translation via ignoring spurious
  correlations.
\newblock In {\em ACL}.

\bibitem[\protect\citeauthoryear{Hassan \bgroup et al\mbox.\egroup
  }{2018}]{hassan2018achieving}
Hassan, H.; Aue, A.; Chen, C.; Chowdhary, V.; Clark, J.~R.; Federmann, C.;
  Huang, X.; Junczys-Dowmunt, M.; Lewis, W.; Li, M.; Liu, S.; Liu, T.~M.; Luo,
  R.; Menezes, A.; Qin, T.; Seide, F.; Tan, X.; Tian, F.; Wu, L.; Wu, S.; Xia,
  Y.; Zhang, D.; Zhang, Z.; and Zhou, M.
\newblock 2018.
\newblock Achieving human parity on automatic chinese to english news
  translation.
\newblock {\em ArXiv} abs/1803.05567.

\bibitem[\protect\citeauthoryear{Howard and
  Ruder}{2018}]{Howard2018UniversalLM}
Howard, J., and Ruder, S.
\newblock 2018.
\newblock Universal language model fine-tuning for text classification.
\newblock In {\em ACL}.

\bibitem[\protect\citeauthoryear{Huang \bgroup et al\mbox.\egroup
  }{2019}]{Huang2019UnicoderAU}
Huang, H.; Liang, Y.; Duan, N.; Gong, M.; Shou, L.; Jiang, D.; and Zhou, M.
\newblock 2019.
\newblock Unicoder: A universal language encoder by pre-training with multiple
  cross-lingual tasks.
\newblock {\em ArXiv} abs/1909.00964.

\bibitem[\protect\citeauthoryear{Johnson \bgroup et al\mbox.\egroup
  }{2016}]{johnson2017google}
Johnson, M.; Schuster, M.; Le, Q.~V.; Krikun, M.; Wu, Y.; Chen, Z.; Thorat, N.;
  Vi{\'e}gas, F.~B.; Wattenberg, M.; Corrado, G.~S.; Hughes, M.; and Dean, J.
\newblock 2016.
\newblock Google’s multilingual neural machine translation system: Enabling
  zero-shot translation.
\newblock {\em Transactions of the Association for Computational Linguistics}
  5:339--351.

\bibitem[\protect\citeauthoryear{Kauers \bgroup et al\mbox.\egroup
  }{2002}]{kauers2002interlingua}
Kauers, M.; Vogel, S.; F{\"u}gen, C.; and Waibel, A.~H.
\newblock 2002.
\newblock Interlingua based statistical machine translation.
\newblock In {\em INTERSPEECH}.

\bibitem[\protect\citeauthoryear{Kim \bgroup et al\mbox.\egroup
  }{2019}]{Kim2019PivotbasedTL}
Kim, Y.; Petrov, P.; Petrushkov, P.; Khadivi, S.; and Ney, H.
\newblock 2019.
\newblock Pivot-based transfer learning for neural machine translation between
  non-english languages.
\newblock {\em ArXiv} abs/1909.09524.

\bibitem[\protect\citeauthoryear{Kim, Gao, and Ney}{2019}]{kim2019effective}
Kim, Y.; Gao, Y.; and Ney, H.
\newblock 2019.
\newblock Effective cross-lingual transfer of neural machine translation models
  without shared vocabularies.
\newblock In {\em ACL}.

\bibitem[\protect\citeauthoryear{Kocmi and Bojar}{2018}]{kocmi2018trivial}
Kocmi, T., and Bojar, O.
\newblock 2018.
\newblock Trivial transfer learning for low-resource neural machine
  translation.
\newblock In {\em WMT}.

\bibitem[\protect\citeauthoryear{Koehn and Knowles}{2017}]{koehn2017six}
Koehn, P., and Knowles, R.
\newblock 2017.
\newblock Six challenges for neural machine translation.
\newblock In {\em NMT@ACL}.

\bibitem[\protect\citeauthoryear{Koehn}{2005}]{koehn2005europarl}
Koehn, P.
\newblock 2005.
\newblock Europarl: A parallel corpus for statistical machine translation.

\bibitem[\protect\citeauthoryear{Lample and Conneau}{2019}]{lample2019cross}
Lample, G., and Conneau, A.
\newblock 2019.
\newblock Cross-lingual language model pretraining.
\newblock {\em ArXiv} abs/1901.07291.

\bibitem[\protect\citeauthoryear{Lample \bgroup et al\mbox.\egroup
  }{2018}]{lample2018phrase}
Lample, G.; Ott, M.; Conneau, A.; Denoyer, L.; and Ranzato, M.
\newblock 2018.
\newblock Phrase-based \& neural unsupervised machine translation.
\newblock In {\em EMNLP}.

\bibitem[\protect\citeauthoryear{Nguyen and Chiang}{2017}]{nguyen2017transfer}
Nguyen, T.~Q., and Chiang, D.
\newblock 2017.
\newblock Transfer learning across low-resource, related languages for neural
  machine translation.
\newblock In {\em IJCNLP}.

\bibitem[\protect\citeauthoryear{Pires, Schlinger, and
  Garrette}{2019}]{DBLP:journals/corr/abs-1906-01502}
Pires, T.; Schlinger, E.; and Garrette, D.
\newblock 2019.
\newblock How multilingual is multilingual bert?
\newblock In {\em ACL}.

\bibitem[\protect\citeauthoryear{Ren \bgroup et al\mbox.\egroup
  }{2019}]{ren2019unsupervised}
Ren, S.; Zhang, Z.; Liu, S.; Zhou, M.; and Ma, S.
\newblock 2019.
\newblock Unsupervised neural machine translation with smt as posterior
  regularization.
\newblock In {\em AAAI}.

\bibitem[\protect\citeauthoryear{Sennrich, Haddow, and
  Birch}{2015}]{sennrich2015neural}
Sennrich, R.; Haddow, B.; and Birch, A.
\newblock 2015.
\newblock Neural machine translation of rare words with subword units.
\newblock In {\em ACL}.

\bibitem[\protect\citeauthoryear{Utiyama and
  Isahara}{2007}]{utiyama2007comparison}
Utiyama, M., and Isahara, H.
\newblock 2007.
\newblock A comparison of pivot methods for phrase-based statistical machine
  translation.
\newblock In {\em HLT-NAACL}.

\bibitem[\protect\citeauthoryear{Vaswani \bgroup et al\mbox.\egroup
  }{2017}]{vaswani2017attention}
Vaswani, A.; Shazeer, N.; Parmar, N.; Uszkoreit, J.; Jones, L.; Gomez, A.~N.;
  Kaiser, L.; and Polosukhin, I.
\newblock 2017.
\newblock Attention is all you need.
\newblock In {\em NIPS}.

\bibitem[\protect\citeauthoryear{Wu \bgroup et al\mbox.\egroup
  }{2016}]{wu2016google}
Wu, Y.; Schuster, M.; Chen, Z.; Le, Q.~V.; Norouzi, M.; Macherey, W.; Krikun,
  M.; Cao, Y.; Gao, Q.; Macherey, K.; et~al.
\newblock 2016.
\newblock Google's neural machine translation system: Bridging the gap between
  human and machine translation.
\newblock {\em arXiv preprint arXiv:1609.08144}.

\bibitem[\protect\citeauthoryear{Zheng, Cheng, and
  Liu}{2017}]{zheng2017maximum}
Zheng, H.; Cheng, Y.; and Liu, Y.~P.
\newblock 2017.
\newblock Maximum expected likelihood estimation for zero-resource neural
  machine translation.
\newblock In {\em IJCAI}.

\bibitem[\protect\citeauthoryear{Zhu \bgroup et al\mbox.\egroup
  }{2013}]{zhu2013improving}
Zhu, X.; He, Z.; Wu, H.; Wang, H.; Zhu, C.; and Zhao, T.
\newblock 2013.
\newblock Improving pivot-based statistical machine translation using random
  walk.
\newblock In {\em EMNLP}.

\bibitem[\protect\citeauthoryear{Zoph \bgroup et al\mbox.\egroup
  }{2016}]{zoph2016transfer}
Zoph, B.; Yuret, D.; May, J.; and Knight, K.
\newblock 2016.
\newblock Transfer learning for low-resource neural machine translation.
\newblock In {\em EMNLP}.

\end{thebibliography}

\end{document}